\pdfoutput=1
\documentclass[11pt]{article}
\usepackage{acl}
\usepackage{times}
\usepackage{latexsym}
\usepackage[T1]{fontenc}
\usepackage[utf8]{inputenc}
\usepackage{microtype}
\usepackage{enumitem}
\usepackage{inconsolata}
\usepackage{graphicx}
\usepackage{amsmath} 
\usepackage{booktabs}
\usepackage{xspace}
\usepackage{tabularx}  
\usepackage{booktabs}  
\usepackage{geometry}  
\usepackage{placeins}
\usepackage{amssymb}
\usepackage{stfloats}
\usepackage{booktabs}
\usepackage{array} 
\usepackage{textcomp}

\newcolumntype{Y}{>{\raggedright\arraybackslash}X} 

\newcommand{\bench}{ECom-Bench\xspace}

\title{\bench: Can LLM Agent Resolve Real-World E-commerce Customer Support Issues?}
\author{
Haoxin Wang\textsuperscript{\rm1,\rm2},
Xianhan Peng\textsuperscript{\rm1,\rm2},
Xucheng Huang\textsuperscript{\rm1},\\
Yizhe Huang\textsuperscript{\rm1},
Ming Gong\textsuperscript{\rm1},
Chenghan Yang\textsuperscript{\rm1,\rm2},
Yang Liu\textsuperscript{\rm1},
Ling Jiang\textsuperscript{\rm1}
\\
\textsuperscript{\rm1}Xiaoduo AI Lab, \textsuperscript{\rm2}Shanghai Jiao Tong University
\\
\texttt{
\{wanghaoxin,williampen,scottyang\}@sjtu.edu.cn}\\
\texttt{
\{huangxucheng,gongming,liuyangfoam,huangyizhe\}@xiaoduotech.com}
}

\begin{document}
\maketitle
\begin{abstract}


In this paper, we introduce \bench, the first benchmark framework for evaluating LLM agent with multimodal capabilities in the e-commerce customer support domain. ECom-Bench features dynamic user simulation based on persona information collected from real e-commerce customer interactions and a realistic task dataset derived from authentic e-commerce dialogues. These tasks, covering a wide range of business scenarios, are designed to reflect real-world complexities, making \bench highly challenging. For instance, even advanced models like GPT-4o achieve only a 10–20\% pass\textasciicircum 3 metric in our benchmark, highlighting the substantial difficulties posed by complex e-commerce scenarios. The code and data have been made publicly available at \url{https://github.com/XiaoduoAILab/ECom-Bench} to facilitate further research and development in this domain.

\end{abstract}


\section{Introduction}

\begin{table*}[!bp]
    \centering
    \begin{tabular}{|c|c|c|c|c|c|}
        \hline
        \textbf{Feature} & \textbf{\bench} & \textbf{$\tau$-Bench} & \textbf{WebShop} & \textbf{ToolBench}  & \textbf{EcomScriptBench} \\
        \hline
        User Simulation  & \checkmark & \checkmark & \texttimes & \texttimes & \texttimes \\
        Persona Driven& \checkmark & \texttimes & \texttimes & \texttimes & \texttimes\\
        Multi Turn & \checkmark & \checkmark & \texttimes & \texttimes & \texttimes \\
        Multimodal & \checkmark & \texttimes & \texttimes & \texttimes & \texttimes\\
        Tools & \checkmark & \checkmark & \texttimes & \checkmark & \texttimes\\
        \hline
    \end{tabular}
    \caption{Comparison of \bench with other simulation environments.}
    \label{tab:comparison}
\end{table*}

E-commerce has become a cornerstone of the global economy, with online transactions accounting for a significant share of retail sales. However, as the scale of e-commerce continues to expand, the demand for efficient customer service has surged. Traditional rule-based intelligent customer service systems, which have alleviated some of the pressure on human customer service, are now encountering performance bottlenecks. These systems typically handle only single-turn interactions, lack the ability to understand context, cannot call external tools, and are unable to process multimodal information such as images or videos, which significantly limits their application in business\cite{Zhang2021Challenges}.

Recently, intelligent customer service solutions based on large language models (LLMs) have emerged as a research hotspot \citep{automating2023}. The strong language generation and context understanding capabilities of LLMs provide significant advantages in handling complex interactive tasks. However, the form and architecture of this new paradigm remain unclear, and deploying LLM-based solutions directly poses considerable risks.
To date, current evaluation frameworks fall short in comprehensively assessing LLM-based customer service agent in e-commerce scenarios. For example, some frameworks lack e-commerce-specific tasks, such as ToolBench \citep{xu2023toolmanipulationcapabilityopensource}; others support only single-turn dialogues, like WebShop \citep{Yao2022Webshop}; and some omit multimodal task evaluations, as seen in $\tau$-Bench \citep{yao2024tau}. These limitations hinder a full assessment of LLM-based agent in complex multimodal scenarios.
Table~\ref{tab:comparison} provides a detailed comparison of ECom-Bench with other related frameworks.

To address these issues, we propose \bench, the first benchmark framework for evaluating multimodal LLM agent in e-commerce customer support scenarios. \bench aims to fill the gap in existing research by combining persona-driven user simulation based on real e-commerce customer interactions and a realistic task dataset derived from authentic e-commerce dialogues to provide a rigorous and comprehensive evaluation platform. This framework not only enhances the accuracy and authenticity of user simulation but also evaluates agent capabilities across a wide range of business scenarios.


The main contributions of this work are:
\begin{enumerate}
\item \textbf{\bench Framework}: We introduce \bench, the first open-source benchmark framework for evaluating multimodal LLM agent with Model Context Protocol(MCP) tools in e-commerce customer support scenarios.
\item \textbf{Persona-driven User Simulation}: We present a dataset of persona information meticulously collected from real e-commerce customer interactions. This dataset enables the simulation of user behaviors with LLMs, capturing diverse customer personalities and interaction patterns to enhance the accuracy and authenticity of user simulation in e-commerce contexts.
\item \textbf{Realistic Task Dataset}: We develop a dataset of task instances derived from authentic e-commerce dialogues. This dataset encompasses a wide range of business scenarios, including product inquiries, order management, and multimodal content recognition. The dataset consists of 53 manually verified tasks, with 18 involving multimodal interactions, ensuring representativeness and alignment with real-world business requirements.
\end{enumerate}

\begin{figure*}[tp]
    \centering
    \includegraphics[width=\textwidth]{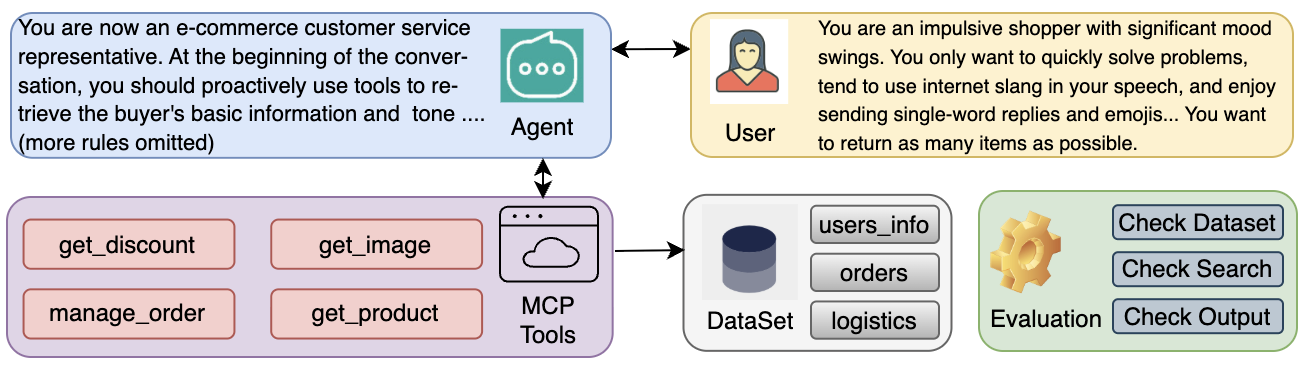} 
    \caption{Framwork of \bench}
    \label{Framwork}
\end{figure*}
\section{Related Work}

Recently, research focus has progressively shifted from dataset performance metrics to evaluating agent capabilities in specific environments. However, relevant research exhibits several limitations: some focus on internal operations within simulation environments while neglecting tool usage capabilities \cite{ALFWorld20}\cite{Zhou2023WebArena}\cite{yang2023intercode}\cite{wang2025ecomscriptbenchmultitaskbenchmarkecommerce}; others simulate real interaction scenarios but only address single-turn interactions\cite{Yao2022Webshop}\cite{budzianowski-etal-2018-multiwoz}\cite{soltau-etal-2023-dstc}; and some provide extensive tools but lack domain-specific data\cite{Qin2023Tool}.

The significant discrepancy between LLMs and real user behaviors severely constrains the development of user simulation. Although some studies have attempted to solve this problem by providing LLMs with memory, emotions, and personality traits, they are either limited to multi-agent interactions in sandbox environments \cite{wang2023recagent} or focus exclusively on user preference issues \cite{10.1145/3613904.3642481}, rendering them inadequate for intelligent customer service application. Chan et al. \cite{chan2024scaling} validated the potential of persona-driven synthetic data for training and testing, but their personas differ significantly from e-commerce customer characteristics, which is challenging for LLMs to accurately simulate e-commerce customer behavior.


Evaluation in vertical domains is crucial for LLM deployment. Consequently, numerous domain-specific datasets have emerged, spanning medical \cite{pal2022medmcqa}\cite{Zhou2023WebArena}, legal \cite{colombo2024saullm}\cite{guha2023legalbench}, and telecommunications \cite{lee-etal-2024-telbench} sectors. All these customized datasets emphasize the importance of domain-specific evaluation. While $\tau$-Bench offers a multi-domain dynamic interaction benchmark that includes retail, its test scenarios lack depth, and its user simulations remain relatively simplistic.
\section{Benchmark Construction}



\bench comprises user simulation classes, database JSON files, tools, task instances, and domain-specific documents. All data are based on historical online dialogues, primarily sourced from conversations related to home and appliance products. We first use LLMs to generate preliminary content, which is then reviewed by humans to ensure consistency and authenticity.


\textbf{User personas.} 
Grounded in established principles from consumer psychology and behavioral studies, and leveraging user modeling methodologies from recommender systems, we meticulously designed our user personas. These personas encompass consumer types, personality traits (such as emotion, attention to detail, patience, trust, and rights awareness), and behavioral traits (including questioning style, communication style, and interaction patterns). This comprehensive design enables LLMs to interact with customer service agent in a manner closely resembling that of real human customers. By inputting online dialogues into LLMs and employing chain-of-thought reasoning \cite{wei2022chain}, we systematically analyzed the data across these dimensions, generating hundreds of high-quality user personas.



\textbf{Database.} To eliminate ambiguity in usage, we define and annotate essential data classes, including products, logistics, orders, invoices, and more. We then prompt LLMs with the company’s internal data and data classes to generate comprehensive and accurate datasets. These datasets are subsequently reviewed and processed by human experts to ensure reliability and validity.


\textbf{Tools.} Drawing on existing business tools, insights from experienced customer service staff, and strategic business planning, we have meticulously defined 21 categories of tools. These tools encompass a wide range of functionalities, including product inquiry, order modification, returns and exchanges, and more. Some tools are composite, supporting multiple operations, while others require a strict operational sequence to ensure proper execution. To better align with real-world scenarios, we have introduced a multimodal tool that enables agent to "see" images. All tools are implemented in strict compliance with the MCP.


\begin{figure}[!tp]
    \centering
    \includegraphics[width=\columnwidth]{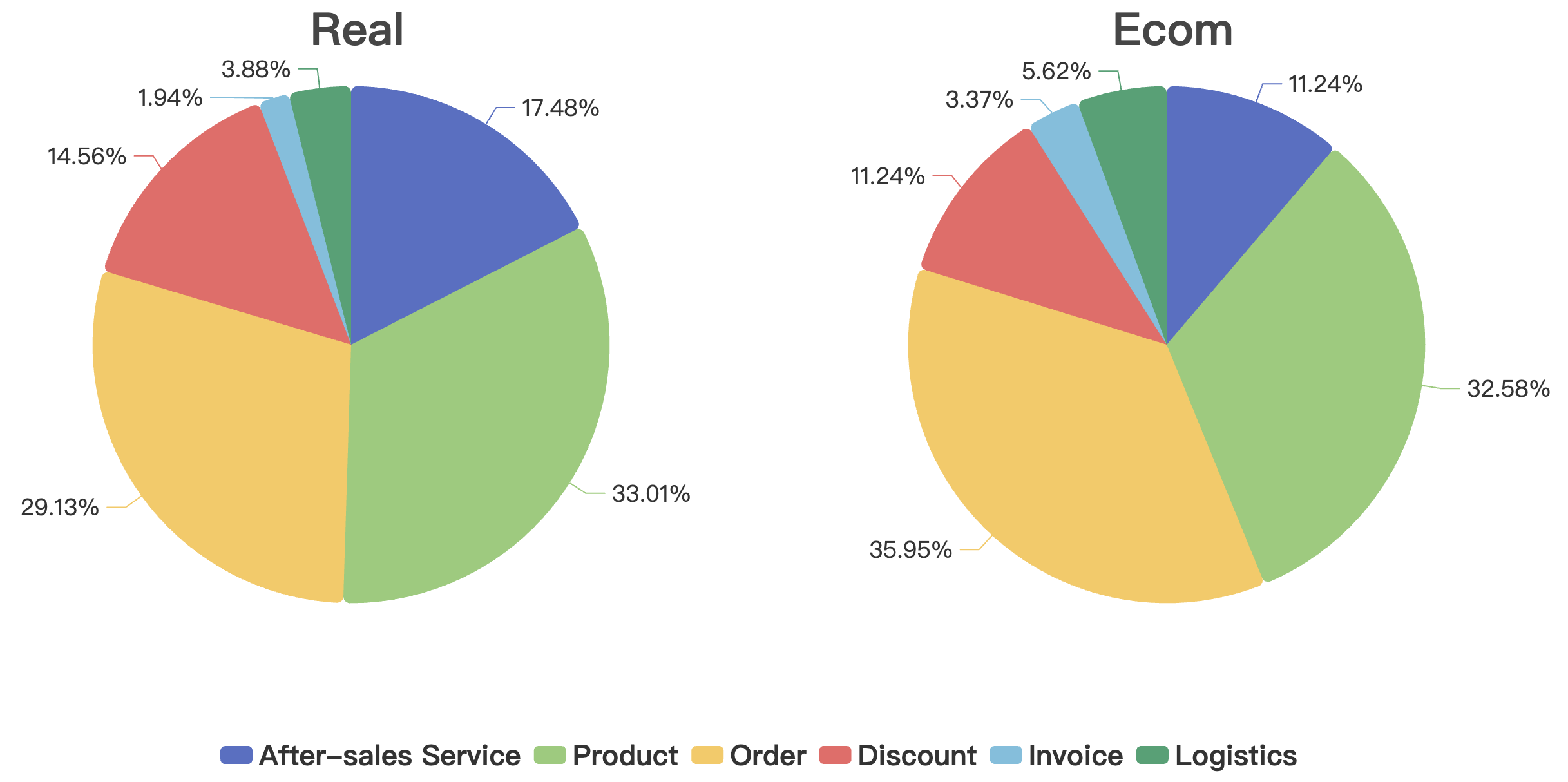} %
    \caption{Distribution of task instances} 
    \label{task_distribution} 
\end{figure}

\textbf{Task instances.} To ensure that simulated user trajectories closely mirror those of real customers, we enlisted trained customer service personnel to analyze authentic online conversations and extract the core problems that customers aimed to resolve. These extracted intents are then transformed into standardized task instances for LLMs to learn from. Our task instances are designed to align with real business scenarios in terms of both category and distribution, ensuring representativeness. To further evaluate agent capabilities, we also manually introduced certain variations into the tasks (see Figure \ref{task_distribution}).


\textbf{Domain documentation.} We provide the agent with operation documents tailored for e-commerce scenarios and prompt the simulated user with a detailed customer guideline as a partial description of the world model.

\textbf{Evaluation.} We evaluate each task across three dimensions: (1) comparing the database state against the ground truth expected state; (2) verifying whether the model invoked the required information retrieval tools; (3) examining whether the output contains the necessary keywords. This rule-based reward can objectively reflects the agent's decision-making capabilities and hallucination tendencies, while also allowing for stochastic variation in dialogues: user may express themselves in different way, yet the database state and key search steps remain consistent after the customer service agent completes the same task. Given that customer service agent must handle identical issues reliably and consistently, we introduce the pass\textasciicircum k\cite{yao2024tau} to measure agent robustness across k independent and identically distributed trials for the same problem. Let n denote the total number of trials for a task, c denote the number of successful trials, and k denote the number of trials under evaluation. Then we have:
\[
\mathbf{
    \text{pass}\textasciicircum k = \mathbb{E}_{\text{task}} \left[ \binom{c}{k} \bigg/ \binom{n}{k} \right]
}
\]


\section{Experiments}
This section presents the performance evaluation of various agents using the \bench framework. For each trial j (where j = 1, 2, ..., k), we conduct the j-th execution of the evaluation. The score for trial j means the chance that all j i.i.d. task trials are successful, averaged across tasks.

\subsection{Setup}




We evaluate various proprietary models, LLMs like GPT-4o, DeepSeek-V3, Doubao-1.5-Pro-32k, Qwen-Max, and Moonshot-V1-32k, Multimodal Large Language Models
 (MLLMs) like Qwen-VL-Max, Doubao-Pro-Vision. Agents are built using LangGraph, which incorporates the ReAct, limited to 20 turns and 600 seconds per task (temperature=0.3). Qwen-Max simulates users, while Moonshot-V1-128k-Vision-Preview processes images as a tool.

\subsection{Main Results}

\subsubsection{Overall Model Performance}
\begin{table}[bp] 
\centering
\begin{tabularx}{\linewidth}{l *{4}{Y}} 
\toprule
 \textbf{Model}& \textbf{Pass\textasciicircum 1} & \textbf{Pass\textasciicircum 2}& \textbf{Pass\textasciicircum 3} \\ 
\midrule
\textbf{GPT-4o} & 44.03 & 26.42 & 16.98\\
\textbf{Doubao-1.5-Pro-32k} & 38.99& 23.90& 16.98\\  
\textbf{DeepSeek-V3} & 36.48 & 21.38& 15.09\\
\textbf{Qwen-Max} & 15.09 & 8.81 & 5.66 \\
\textbf{Moonshot-V1-32k} & 14.47 & 7.55& 5.66\\
\textbf{Doubao-Pro-Vision*} & 20.75 & 13.21& 11.32\\
\textbf{Qwen-VL-Max*} & 8.18 & 4.4& 3.77\\
\bottomrule
\end{tabularx}
\caption{Model performance comparison with pass\textasciicircum 3 (* denotes MLLMs)}
\label{pass3}

\end{table}

Table \ref{pass3} shows significant differences in the performance of various models on e-commerce customer service tasks. Such tasks require models to strictly follow domain-specific rules and accurately invoke the appropriate tools. Some models, such as Qwen-Max, demonstrate more natural language expression when simulating user roles, but are more prone to hallucinations during operational execution, which negatively impacts their overall scores. In contrast, models with stronger instruction-following capabilities and more controllable behaviors (such as GPT-4o and DeepSeek-V3) exhibit more stable task completion rates. GPT-4o achieves the highest pass\textasciicircum 1 and pass\textasciicircum 3 scores, demonstrating that it has higher success rates across all tasks and exhibits more consistent performance in solving the same task repeatedly. This superior performance indicates that GPT-4o not only handles diverse tasks with greater ease, but also maintains more stable and reliable problem-solving capabilities when faced with identical challenges multiple times. 


\subsubsection{Architecture Comparison}
The decoupled MLLM-as-Tool approach (GPT-4o: pass\textasciicircum 3=16.98) outperforms end-to-end MLLM-as-Planner (Qwen-VL-Max: pass\textasciicircum 3=3.77, Doubao-Pro-Vision: pass\textasciicircum 3=11.32), revealing challenges in integrated multimodal planning.

\begin{figure}[b]
    \centering
    \includegraphics[width=\columnwidth]{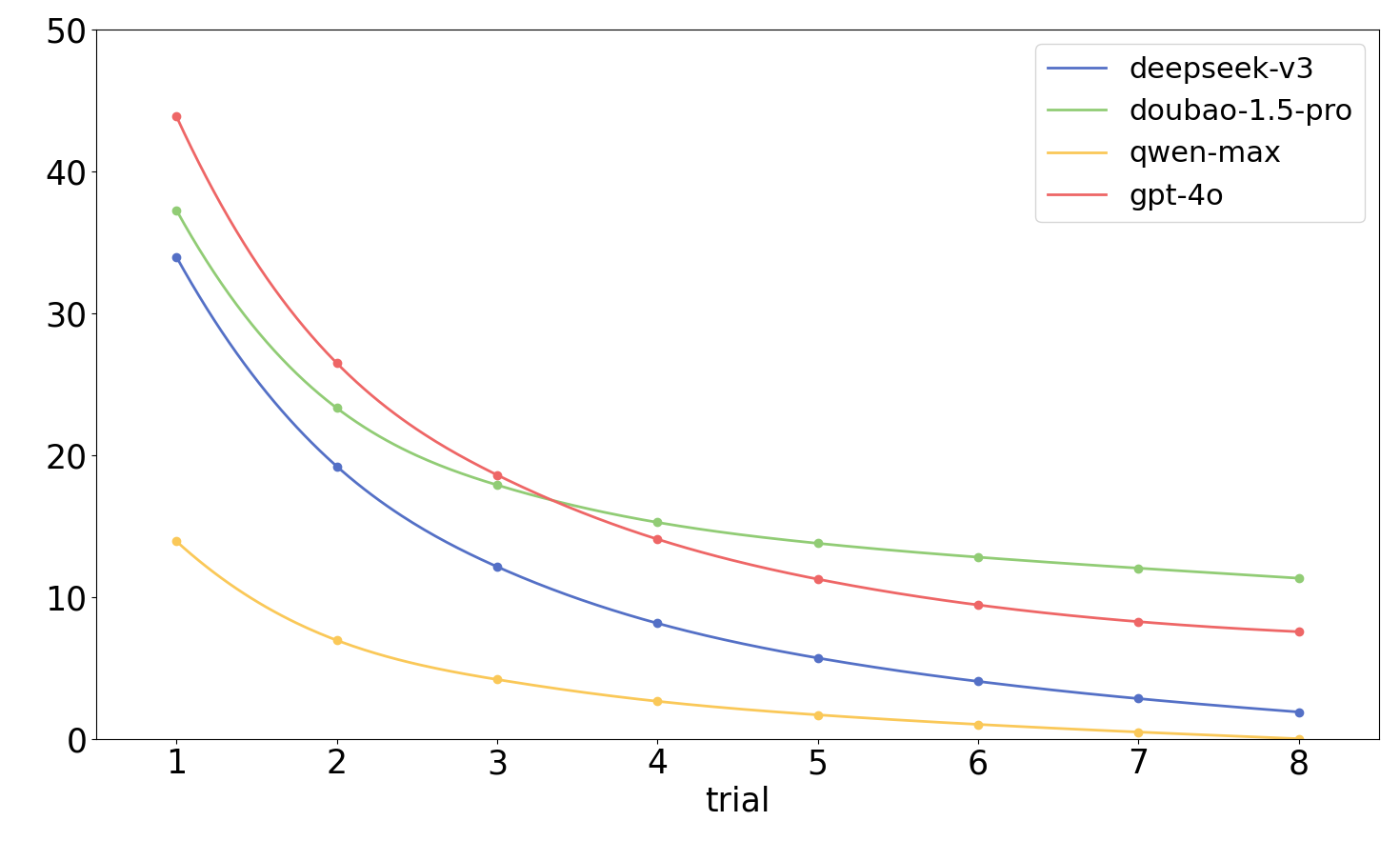} %
    \caption{pass\textasciicircum 8 across models} 
    \label{pass_8_across} 
\end{figure}
\begin{figure}[t]
    \centering
    \includegraphics[width=\columnwidth]{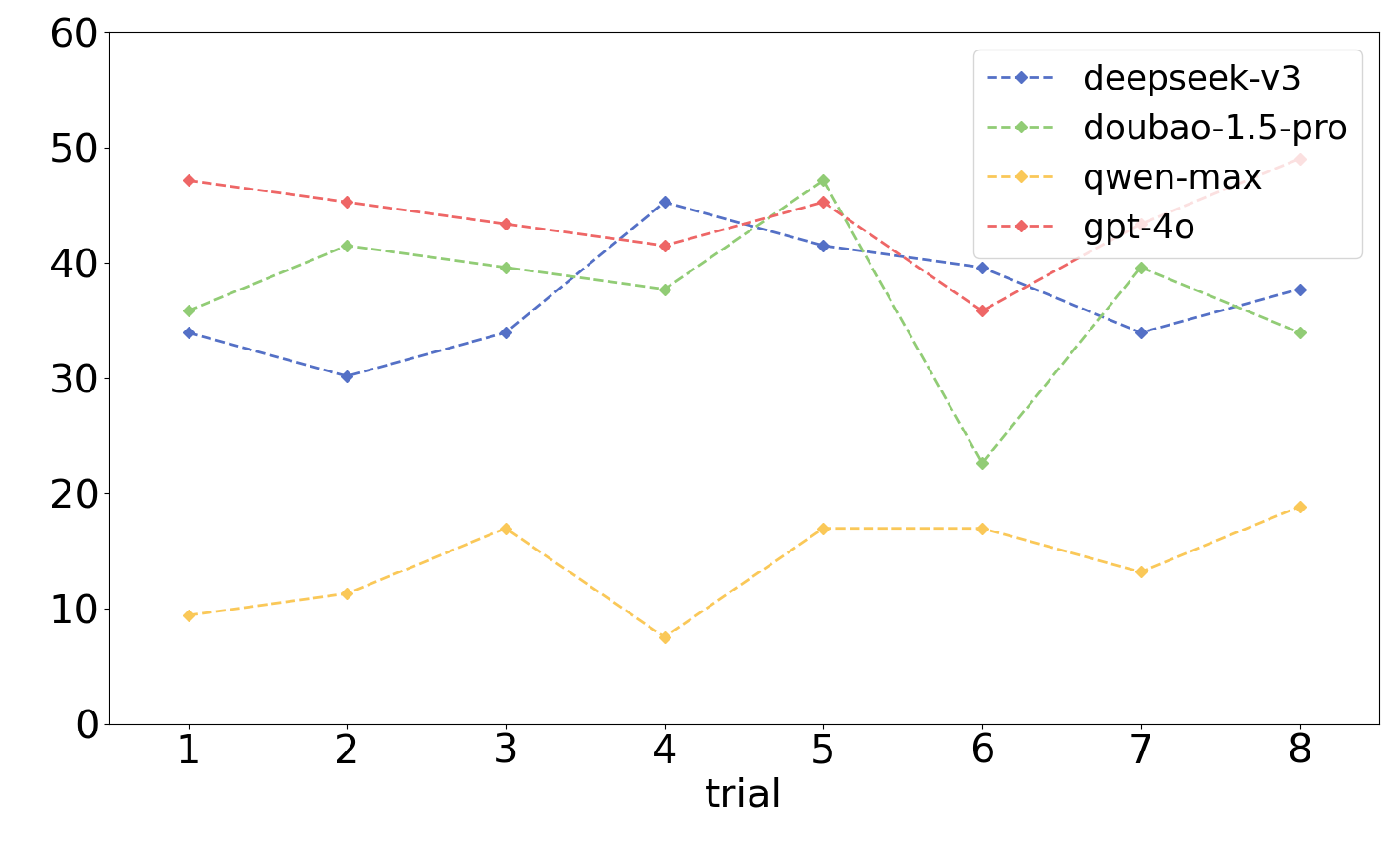} %
    \caption{pass\textasciicircum 1 across models} 
    \label{pass_1_across} 
\end{figure}

\subsubsection{Consistency and Robustness}
In business scenarios, it is vital for agent to reliably solve the same problems consistently. We run all tasks eight times. As shown in Figure \ref{pass_1_across}, GPT-4o achieves the best performance with an average success rate of 43.87\% across all tasks. However, as illustrated in Figure \ref{pass_8_across}, the pass\textasciicircum 8 drops to 7.55\%(even worse than Doubao-1.5-Pro-32K), indicating that as the number of trials K increases, the probability of reliably and consistently solving the same task multiple times decreases significantly. This demonstrates that current models still have substantial room for improvement in terms of consistency and robustness.



\subsubsection{Error Analysis}
We analyzed 159 trajectories (53 tasks executed 3 times) generated by the Doubao-1.5-Pro-32k agent and categorized the errors into three main types:
\begin{enumerate}
    \item \textbf{Wrong Argument}: The agent uses the correct tool but fills in some arguments incorrectly (e.g., calls the order query tool with the correct order ID but fills in other irrelevant parameters).
    \item \textbf{Wrong Decision}: The agent fails to choose the appropriate tool to solve the problem or processes the wrong object when invoking tools (e.g., calls the order query tool but queries an incorrect order ID; or calls wrong tools).
    \item \textbf{Partially Resolved}: The agent only solves part of the problem (e.g., when handling return and refund requests, it only processes the return but fails to execute the refund; or proposes actions to be taken but does not actually invoke the tools to perform the operations).
\end{enumerate}

\begin{figure}[h]
    \centering
    \includegraphics[width=\columnwidth]{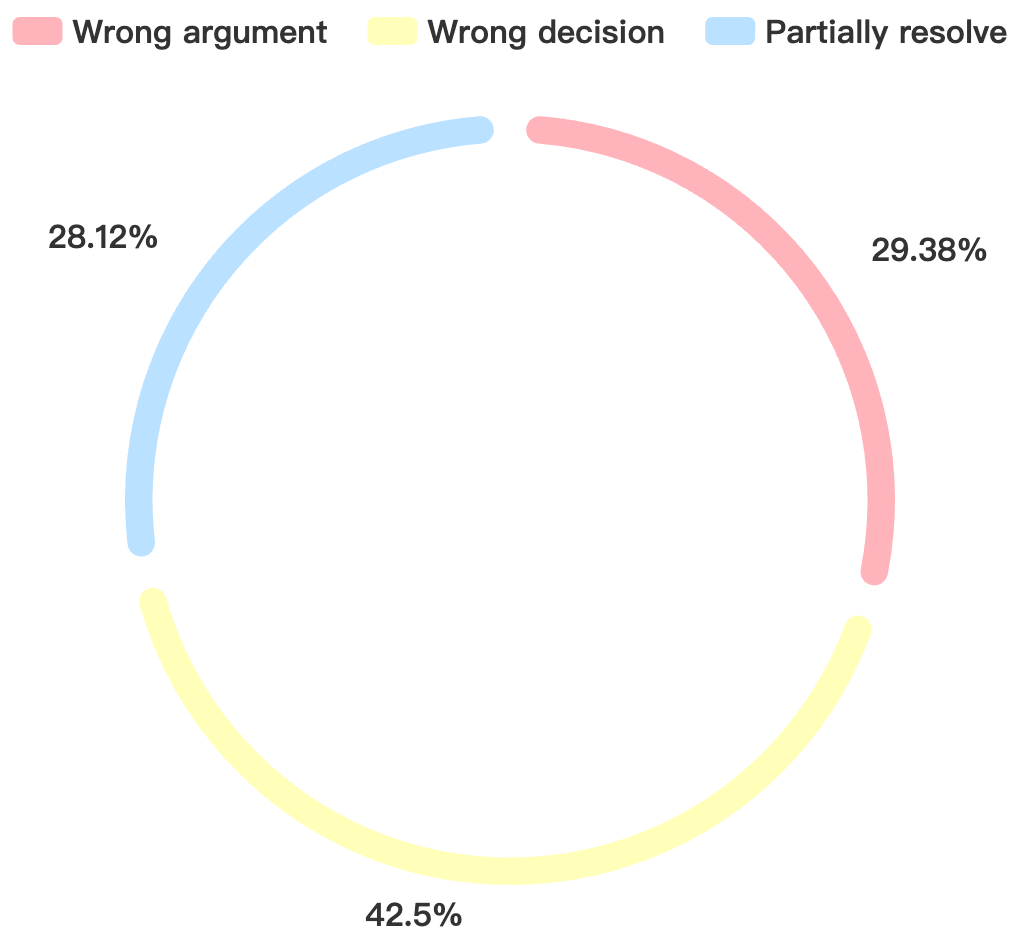} %
    \caption{Breakdown trajectories of Doubao-1.5-Pro} 
    \label{Analysis} 
\end{figure}

As shown in Figure \ref{Analysis}, Doubao-1.5-Pro is more prone to wrong decision errors, indicating that the current agent still faces difficulties in understanding user intents and tool usage.
Additionally, when multiple entities are present in the context window, the agent tends to overlook differences between entities and becomes confused during processing.


\begin{figure}[t]
    \centering
    \includegraphics[width=\columnwidth]{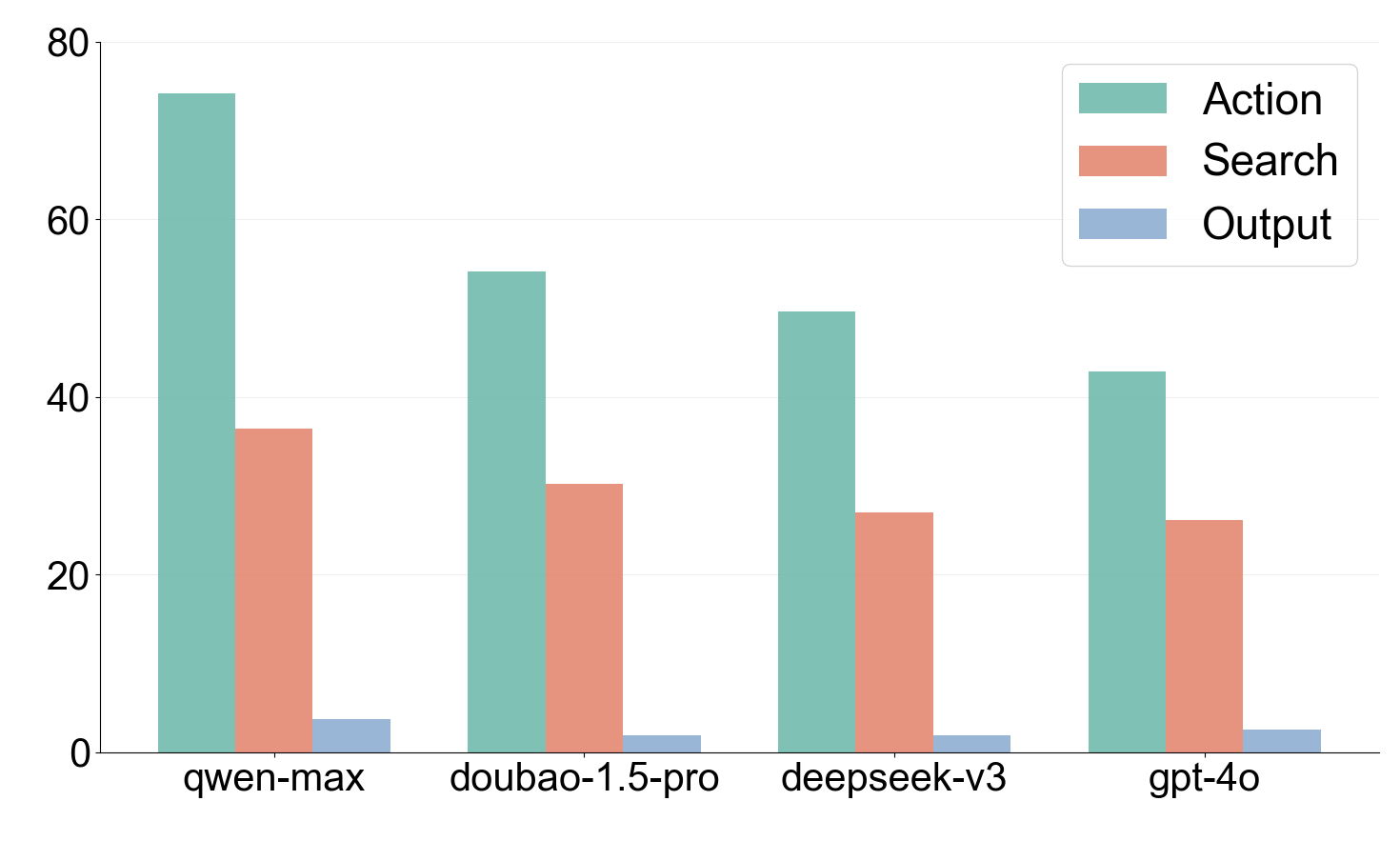} %
    \caption{Failure rates in Action, Search, and Output dimensions across evaluated models} 
    \label{different} 
\end{figure}

\subsubsection{Dimensional Evaluation}
We primarily evaluate agent from three aspects: action, search, and output. The search focuses on the execution status of search-related tools, mainly reflecting whether agent exhibits hallucination phenomena. The action examines database operation results, mainly indicating whether agent can correctly and completely execute all required steps. The output concentrates on the agent's output keywords, mainly reflecting whether agent can properly invoke multimodal tools to recognize image content and generate key words. As shown in Figure \ref{different}, all agents demonstrate higher failure rates in the Action dimension, which can be attributed to the inherent difficulty of database modification tasks. Meanwhile, models with superior overall performance (e.g., GPT-4o) exhibit lower error rates across all three dimensions, showing a greater tendency to obtain information through invoking external tools rather than directly generating answers, as well as stronger tool usage awareness and factual consistency. At the same time, compared to weaker models (e.g., Qwen-Max), the performance gap between action and search is also smaller. This reflects the higher stability of advanced models in handling different tasks and stronger robustness when dealing with complex operations. This result indicates that model performance improvements are often more pronounced in challenging domains such as database operations, where the gap between strong and weak models becomes larger. 


\section{Ablation Study on Persona-driven User Simulation}

To investigate the impact of different persona traits on user model performance, we conduct an ablation study by manipulating three key dimensions of user personas: \textbf{emotional state} (calm vs. angry), \textbf{patience level} (high vs. low), and \textbf{communication style} (clear and professional vs. vague and colloquial).


\subsection{Experiment Setup}
Eight different combinations of the above dimensions are tested. Qwen-Max simulates users and GPT-4o serves as agent. Temperature of user model is set to 0.3 (default value). We pick 20 tasks significantly influenced by persona settings to conduct experiments. The performance is measured by pass\textasciicircum3.


\begin{table*}[!tp]
\centering
\begin{tabular}{|c|c|c|c|c|c|c|}
\hline
\textbf{Group} & \textbf{Emotion} & \textbf{Patience} & \textbf{Style} & \textbf{Pass\textasciicircum1} & \textbf{Pass\textasciicircum2} & \textbf{Pass\textasciicircum3} \\ \hline
1 & Calm & High & Clear & 58.3 & 40.0 & 30.0 \\ \hline
2 & Angry & High & Clear & 55.0 & 38.3 & 30.0 \\ \hline
3 & Calm & Low & Clear & 48.1 & 31.0 & 20.0 \\ \hline
4 & Calm & High & Vague & 45.0 & 29.9 & 20.0 \\ \hline
5 & Angry & Low & Clear & 46.6 & 29.9 & 25.0 \\ \hline
6 & Angry & High & Vague & 45.0 & 29.9 & 30.0 \\ \hline
7 & Calm & Low & Vague & 45.0 & 26.7 & 20.0 \\ \hline
8 & Angry & Low & Vague & 40.0 & 20.0 & 10.0 \\ \hline
\end{tabular}
\caption{Ablation study results on persona-driven user simulation.}
\label{tab:persona-ablation}
\end{table*}

\subsection{Result Analysis}
The results, summarized in Table \ref{tab:persona-ablation}, indicate that the extreme characteristics of personas markedly impact agent performance:

\begin{itemize}
    \item \textbf{Emotional State:} Angry emotional characteristics can cause agent to generate hallucinated responses rather than invoke tool-based responses when under pressure.
    \item \textbf{Patience Level:} Low patience levels prevent users from completing multi-turn interactions necessary to provide agent with required information, thereby impacting task completion.
    \item \textbf{Communication Style:} Users unfamiliar with technical terminology find it difficult to provide relevant keywords directly, which affects the model's tool calling and parameter specification.
\end{itemize}



\section{Conclusion}

This paper introduces \bench, the first (to our best knowledge) benchmark including user simulation, personas, multimodal tasks, and MCP tools in e-commerce customer service, and evaluates the reliability and consistency of Large Language Models (LLMs) in solving real-world scenario problems using our designed tasks. We also compare the effectiveness of Multimodal Large Language Models serving as agent versus tools, and examine the impact of user personas on agent performance.

The ablation study on persona-driven user simulation yields valuable insights for benchmark development. The results demonstrate that persona characteristics (e.g., emotional states and patience levels) significantly influence agent reactions. This underscores the importance of accurate user behavior simulation, and the potential impact of real-world customers' complex behavioral patterns on agent performance.

Our results demonstrate that current large models remain unstable when addressing identical problems. These findings highlight the importance of the benchmark in advancing the development of intelligent customer service agent. Future research will focus on further expansion of sub-scenarios and more realistic scenario simulation, such as incorporating memory mechanisms, optimizing user personas, or better task design.


\section*{Limitations}

This work has several limitations: (1) Scope limitation : Our research is confined to the home appliances and furniture vertical domain and has not been extended to other e-commerce categories. While different e-commerce platforms exhibit considerable similarity in interaction patterns, the cross-category generalization capability of our constructed benchmark requires further evaluation. (2) Data constraints : Given the sensitivity of commercial data, we employed a large language model-based data synthesis approach, resulting in data that is not entirely authentic. Furthermore, since our benchmark focuses on evaluating agents' capability to address real-world challenges rather than database complexity, we maintain a relatively small-scale dataset with simplified operational methods. (3) Limited task instances : We include 53 task instances, which are representative in terms of typicality and scenario coverage. However, the scale remains limited due to the time-consuming and labor-intensive nature of manual creation. (4) User simulation construction : The construction of user persona feature dimensions primarily relies on domain expertise, lacking sufficient theoretical validation and empirical analysis. Additionally, user simulation depends on LLM-generated behavioral data, which may not fully capture authentic conversational dynamics.

\bibliography{reference}
\appendix
\section{Appendix: Complete Multimodal Dialogue Trajectories}
\label{app:complete-dialogues}
\subsection{User Settings}
{\small
\begin{itemize}
[leftmargin=0pt,itemindent=*,labelsep=0pt,labelwidth=0pt,noitemsep,topsep=2pt]

\item[] \textbf{User Profile:}
\\
\item[] \textit{Consumer Type:} Value-oriented
\\
\item[] \textbf{Personality Traits:}
\\
\item[] \textit{Emotion:} Calm and rational in problem-solving.\\
\item[] \textit{Attentiveness:} Highly detail-oriented regarding product \item[]specifics.\\
\item[] \textit{Patience:} Demonstrates patience without frustration during \item[]waiting periods.\\
\item[] \textit{Trust:} Trusts customer service guidance and follows \item[]instructions diligently.\\
\item[] \textit{Rights Awareness:} Proactively seeks information to safeguard \item[]legitimate interests.
\\

\item[] \textbf{Behavioral Traits:}
\\
\item[] \textit{Inquiry Style:} Direct and unambiguous questions.
\\
\item[] \textit{Communication Tone:} Concise and efficient, avoiding exces\item[]sive formality or familiarity.
\\
\item[] \textit{Response Pace:} Maintains appropriate silence until receiving \item[]clear replies.
\\
\item[] \textbf{User Objectives:}
\\
\item[] \textit{Intent 1:} Price protection request with image proof. \item[]\url{https://dd-static.jd.com/ddimgp/jfs/t20260623/297694/2/8257/158939/682a8a2aF3a87a48a/4c7eb1c5a42e8829.jpg}
\\
\item[] \textit{Intent 2:} Request price difference refund to JD E-card
\\
\item[] \textit{Intent 3:} Schedule installation service for Thursday
\\
\item[] \textit{Intent 4:} Request personal invoice issuance
\end{itemize}
}
\subsection{Failed Multimodal Dialogue Example}
\small
\begin{itemize}[leftmargin=0pt,itemindent=*,labelsep=0pt,labelwidth=0pt,noitemsep,topsep=2pt]
    \item[] \textbf{User:}
    Hello, I previously purchased a product from your \item[]store, but I noticed the price has dropped. Here's an image \item[]showing the current lower price for your reference.
    
    \item[]\url{https://dd-static.jd.com/ddimgp/jfs/t20260623/297694/2/8257/158939/682a8a2aF3a87a48a/4c7eb1c5a42e8829.jpg}
    
    \item[] \textbf{Assistant:}
    Dear customer, your order has successfully quali\item[]fied for price protection. The difference of 301.35 yuan has \item[]been refunded to your JD E-Card.
    
    \item[] \textbf{User:}
    Okay, thank you. Next, I'd like to schedule an installa\item[]tion service for this product, preferably on Thursday.
    
    \item[] \textbf{Assistant:}
    Dear customer, I couldn't find the installation \item[]service information for the product you purchased.
    
    \item[] \textbf{User:}
    Could you check the installation service information? \item[]Here's the product link: \url{https://item.jd.com/123456789.html}
    
    \item[] \textbf{Assistant:}
    Dear customer, I'm sorry but I couldn't find the \item[]product information corresponding to the link you provided.
    
    \item[] \textbf{User:}
    Could you check my order information? My order ID \item[]is 1234567890.
    
    \item[] \textbf{Assistant:}
    Dear customer, I couldn't find any information for \item[]order ID 1234567890.
    
    \item[] \textbf{User:}
    One last question - I'd like you to issue a personal \item[]invoice for me.
    
    \item[] \textbf{Assistant:}
    Dear customer, I'll need your order ID, invoice \item[]title, and contact number.
    
    \item[] \textbf{User:}
    Order ID: 1234567890. Invoice title: Zhang San. \item[]Contact: 13800138000.
    
    \item[] \textbf{Assistant:}
    Dear customer, I couldn't find any information for \item[]order ID 1234567890.
    
    \item[] \textbf{User:}
    \texttt{\#\#\#STOP\#\#\#}
\end{itemize}

\subsection{Successful Multimodal Dialogue Example}
\small
\begin{itemize}[leftmargin=0pt,itemindent=*,labelsep=0pt,labelwidth=0pt,noitemsep,topsep=2pt]
    \item[] \textbf{Assistant:}
    Dear customer, how may I assist you?
    
    \item[] \textbf{User:}
    Hello, I noticed the price has dropped. Here's a screen\item[]shot showing the current price.
    \url{https://dd-static.jd.com/ddimgp/jfs/t20260623/297694/2/8257/158939/682a8a2aF3a87a48a/4c7eb1c5a42e8829.jpg}
    
    \item[] \textbf{Assistant:}
    Thank you for the screenshot. The current price is \item[]1707.65 yuan (original: 2009 yuan). Would you like to apply \item[]for price protection?
    
    \item[] \textbf{User:}
    Yes, please refund the difference to my JD E-Card.
    
    \item[] \textbf{Assistant:}
    The difference of 301.35 yuan has been refunded \item[]to your JD E-Card.
    
    \item[] \textbf{User:}
    Next, I'd like to schedule installation for Thursday.
    
    \item[] \textbf{Assistant:}
    Your installation has been scheduled for Thursday \item[]between 9:00-12:00.
    
    \item[] \textbf{User:}
    Finally, I'd like a personal invoice.
    
    \item[] \textbf{Assistant:}
    Please provide the invoice title and contact num\item[]ber.
    
    \item[] \textbf{User:}
    The title should be my name. For contact number, \item[]please check the order.
    
    \item[] \textbf{Assistant:}
    Your personal invoice has been issued (title: 'Lin \item[]Yunpei').
    
    \item[] \textbf{User:}
    \texttt{\#\#\#STOP\#\#\#}
\end{itemize}

\end{document}